# Understanding food inflation in India: A Machine Learning approach

**Akash Malhotra** [a,1]**, Mayank Maloo**[b]

[a] *Indian Institute of Technology, Bombay, India*
[b] *Indian Institute of Technology, Bombay, India*

# Abstract

*Over the past decade, the stellar growth of Indian economy has been challenged by persistently high levels of inflation, particularly in food prices. The primary reason behind this stubborn food inflation is mismatch in supply-demand, as domestic agricultural production has failed to keep up with rising demand owing to a number of proximate factors. The relative significance of these factors in determining the change in food prices have been analysed using gradient boosted regression trees (BRT) – a machine learning technique. The results from BRT indicates all predictor variables to be fairly significant in explaining the change in food prices, with MSP and farm wages being relatively more important than others. International food prices were found to have limited relevance in explaining the variation in domestic food prices. The challenge of ensuring food and nutritional security for growing Indian population with rising incomes needs to be addressed through resolute policy reforms.*

---

[1] *Corresponding author at:* Indian Institute of Technology, Bombay, Mumbai-400076, India.
Tel.: +91 8828174423.
*E-mail addresses:* akash_malhotra@iitb.ac.in (A. Malhotra), mayank_maloo@iitb.ac.in (M. Maloo).

# 1. Introduction

The second most populous country on the planet, India, has been struggling in recent years to keep its food price inflation[2] within politically acceptable and economically sustainable levels. Unlike developed economies, food inflation has had a significant impact on cumulative inflation, as food expenditure constitutes more than 40 percent[3] of total household expenditure in India. Consequently, future inflation expectations are driven, to a great degree, by food prices in this country - creating a vicious circle. The retail food inflation has grown at an average rate of 9.82 percent since FY07[4] and even crossed double digits in four instances (Fig. 1), with food prices becoming more than double in absolute terms in these last ten years. Barring two years of FY11-12 in which crude oil prices rose rapidly, food inflation has always exceeded the overall inflation in the last decade by more than 2 percentage points.

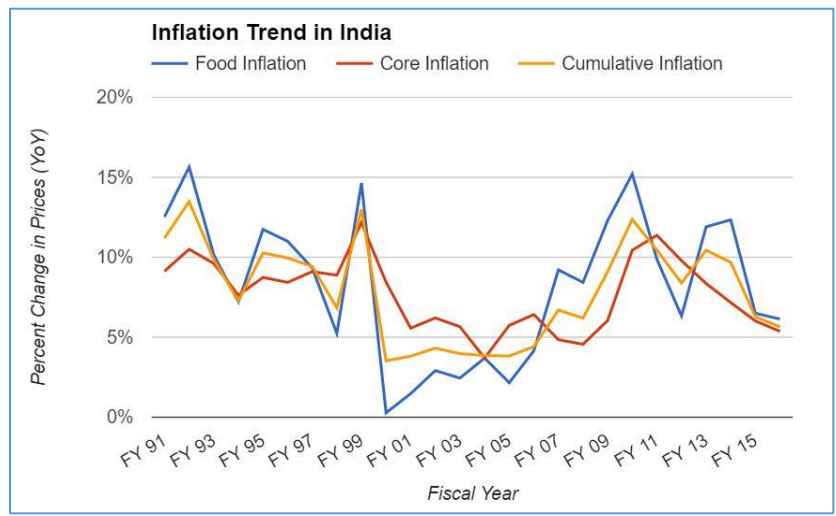

**Figure 1.** Inflation trend based on CPI-IW during FY91-FY16. *Source*: DBIE, RBI

Apart from birthing political scandals, steep rise in food prices creates exigent circumstances for one-fifths of Indian population sustaining below poverty line. Unsustainable rise in food prices inflicts a destructive *'hidden tax'* on poor Indian households which have to spend more than 60 percent on food articles as they generally lack savings and access to financial instruments for hedging against inflation. Such high levels of food inflation are seriously hurting India's fight against poverty and growth prospects which have witnessed a slowdown over the recent past. RBI[5] has repeatedly stressed on the need of containing rise in food prices for effective monetary policy transmission and easing overall headline inflation (RBI 2014; Rajan 2014).

---

[2] For this study, Consumer Price Index (CPI) or retail inflation is based on CPI-IL, unless stated otherwise.
[3] *Source:* 68th NSSO (National Sample Survey Office) Consumption Expenditure Survey 2011-12
[4] FY denotes Fiscal Year; Indian Fiscal Year begins from 1st April and ends on 31st March of next calendar year. For instance, FY07 represents the year starting from April 1, 2006 and ending on March 31, 2007.

With rising incomes and population of India projected to grow at 1.2 percent[6], demand for food articles will continue to increase but the supply has been failing to keep up with rising demand in recent years. It has been noted by previous researchers (Gokarn 2011) that in an equilibrating framework, when food prices rise in the wake of supply stagnation, the most effective way to tackle food inflation is sorting out supply-side factors and ramping up production. The path to achieving food security for growing Indian economy with such a diverse demography is certain to be hindered by push-pull between populistic politics and business interests.

In this backdrop, the current study attempts to review the primary determinants of food inflation in India and statistically analyse their relative significance using a Machine Learning (ML) technique - *Regression with Boosted Decision Trees*. Unlike other domains of science, the adoption of ML in economics has been sparse and slow. ML techniques could potentially serve as a powerful econometric tool in estimating exploratory/predictive economic models on high-dimensional data. However, the usefulness of ML has often been subdued by its limited ability to produce visually interpretable model outcomes. In this context, Boosted Regression Trees (BRT) are particularly promising alternatives because they combine high predictive accuracy with appealing options for the interpretation of model outcomes. In the future, Machine Learning is expected to become a standard part of empirical research in economics as well as contribute to the development of economic theory.

---

[5] RBI (Reserve Bank of India) is the Central bank of India
[6] *Source:* 2015 World Bank estimate

Section 2 presents the context effects and the dynamics of food inflation in India. Section 3 studies the factors driving domestic food prices. Section 4 describes data and statistical model employed and discusses the results from the model. Section 5 concludes with feasible policy recommendations aimed at bringing down food inflation to sustainable levels without adversely affecting growth.

# 2. Country Context and Background

## 2.1 Demographic and Macroeconomic trends

Spurred by wide-ranging economic reforms during 1990s, India has shown exceptional growth in the last decade with its GDP growing at an impressive average rate of 7.5 percent (see Fig. 2). Although share of agriculture in Indian GDP has been in decline (see Fig. 2) owing to expansion in manufacturing and services, the importance of this sector is largely understated by this particular indicator. The significance of agriculture in Indian social and economic fabric is better understood by analysing the rural demographics where two-thirds of India reside. Nearly 70 percent of India's poor live in rural areas where agriculture and allied activities are still the largest source of employment. This makes agriculture, a unique sector dictating both supply and demand of food articles in India. The task of formulating and implementing food policy for more than 1 billion people is challenging in itself, which is complicated further by the fact that about 270 million[7] Indian people are still below the poverty line with an income less than $1.9 a day[8].

---

[7] *Source:* PovcalNet, World Bank; Data last updated on Oct. 1, 2016
[8] International poverty line used by World Bank

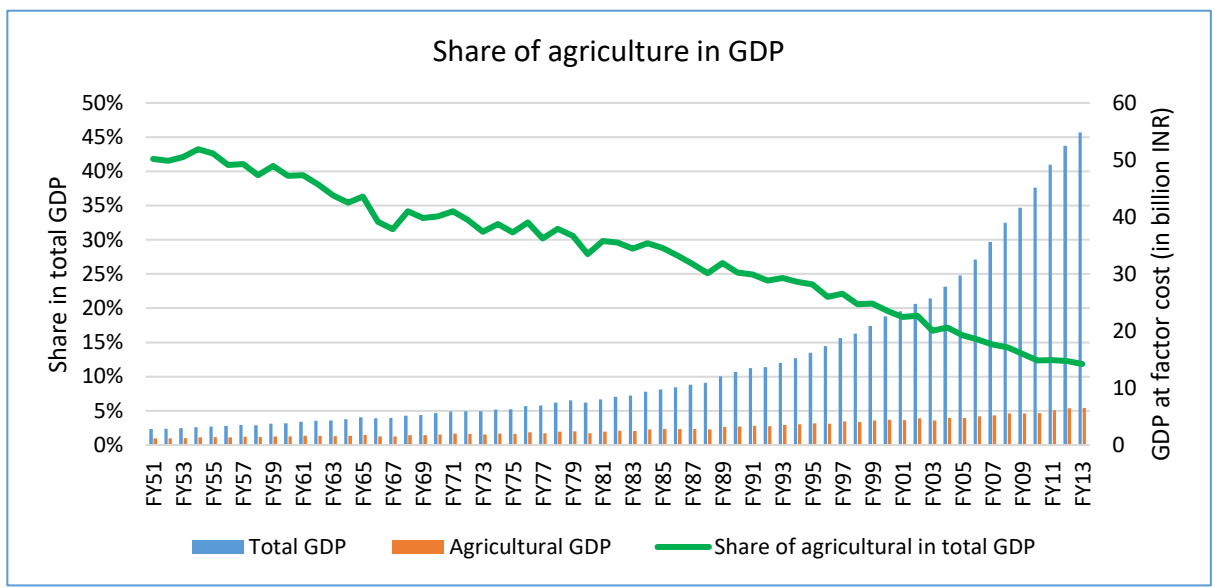

**Figure 2.** GDP at constant prices (2004-05 series). *Source*: National Account Statistics

## 2.2 Food Management Policy

The food management policy in India has primarily focussed on achieving sustainable food security for ever-growing Indian population. This has led to high degree of government involvement with competing objectives such as creating production incentives to farmers, ensuring food security to poor and mitigating effects of supply shocks on prices and farmers arising out of climatic anomalies and global price bouts. After episodes of food crisis in 1970s, India has followed the path of achieving self-sustenance in two main staples - rice and wheat, complemented by centralised procurement of these two crops from the market to meet the needs of buffer stocks and grain distribution system run by central and state governments which delivers rice and wheat to poor consumers at highly subsidised prices. The government exercises control over this policy through twin instruments, viz, Minimum Support Prices (*MSP)* for cultivators and Public Distribution System (*PDS*) for economically weaker sections. Interestingly, there is a significant overlap in these two sections of Indian population as according to a 2014 MOSPI[9] estimate, over 36 percent[10] of agricultural households had qualified for the Below Poverty Line (BPL) ration cards. Where farmers are the major benefactors of both PDS & MSP, it creates baffling push-pull dynamics in policy implementation.

---

[9] MOSPI - Ministry of Statistics and Programme Implementation
[10] Source: 70th NSSO survey on Situation of Agricultural Households in India, December 2013

The current structure of Food Administration in India has been neatly summarised by Saini and Kozicka (2014, pp. 9-14). The MSP for eligible crops are decided and announced at the beginning of sowing season by central government on recommendations of Commission for Agricultural Costs and Prices (CACP). The MSP as a policy instrument, is designed to be the national floor level price at which Food Corporation of India (FCI) procures or buy whatever quantities farmers have to offer. The grain stocks procured through this open-ended operation goes into the central pool maintained by FCI which holds the responsibility of fulfilling buffer stock norms prescribed by central government based on mapping food grain distribution requirements with the food procurement patterns. In addition to maintaining *operational stocks* for various welfare schemes under PDS, FCI maintains a *strategic reserve* to mitigate any future price bouts or unanticipated grain requirements. However, FCI has been repeatedly criticised for holding much more stocks than the prescribed buffer norms and delayed response in releasing stocks during times of scarcity (Chand 2010; Gulati et al. 2012; Anand et al. 2016). This is further aggravated by incremental costs of carrying excess stocks over buffer norms[11].

A look in the history of FCI's procurement policy depicts a counter-cyclical character, setting up inflationary pressures in an already inflated market. Ideally, the FCI is expected to stock up its granaries in times of abundant supplies and release food grains through its open market sale scheme (OMSS) in times of scarcity. However, there have been many instances where FCI has not only withheld stocks during a bad crop year but has also procured more from an already supply-constrained market thereby pushing up prices further. Recently in FY09, when inflation in cereal prices was hovering around 10.5 percent and monsoon rains were below normal, FCI stepped up procurement by almost 30 percent which was followed by an 11 percent rise in cereal prices in the subsequent year (see Fig. 3).

---

[11] Gulati et al. (2012) estimated the combined costs incurred in transporting, storing and distributing food grains to be 50 percent more than procurement prices.

MSPs are declared at the beginning of sowing season, whereas, the actual intake by government to build its buffer stock is done post-harvest, distorting the grain demand-supply equilibrium as it reduces the availability of food grains in open market for regular households. Thus, open market prices are eventually set by combined household and actual post-harvest buffer stock intake. Neglecting international trade, post-harvest short-term supply curve for food grains could be assumed fixed or vertical as depicted in Fig. 4. The government demand curve could also be assumed as vertical, as historical data reveals that price levels does not usually affect FCI's decision regarding buffer stock intake. This implies that any increase in buffer stock intake level causes a rightward parallel shift of the combined demand curve of household and government, thereby reducing the availability in open market for households which ultimately, raises the open market prices.

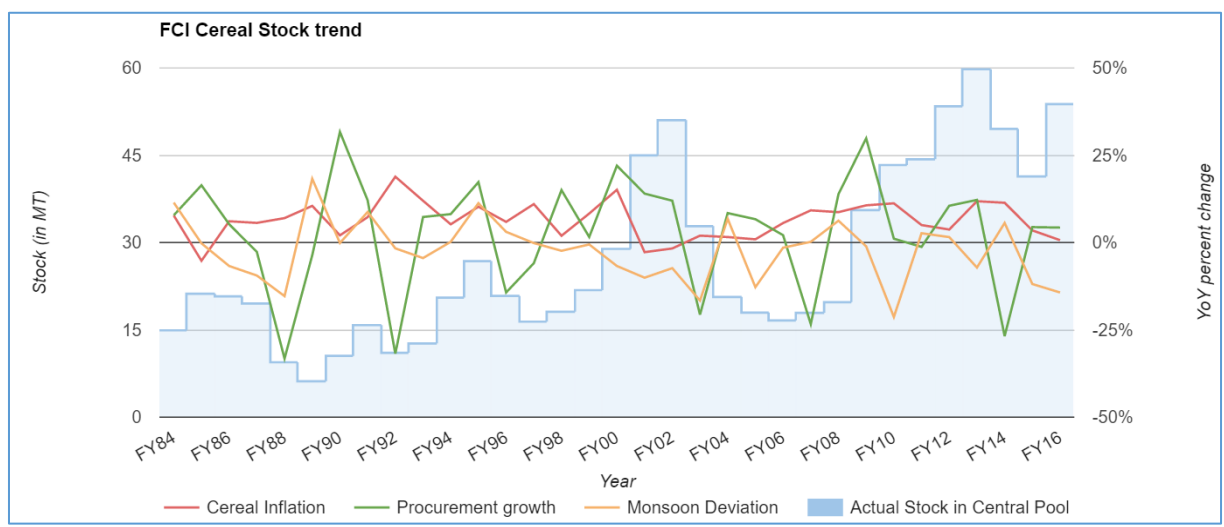

**Figure 3.** Cereal PDS supply management (*Data Source*: DBIE, RBI)

In nutshell, the buffer stocking policy of food grains practiced by India is conflicting in itself. Using the same set of instruments to incentivise agricultural production by ensuring remunerative prices to farmers, mitigating volatility in grain prices and providing subsidised food security to poor at the same time creates many inefficiencies and leakages[12] in the system along with a huge spread between the purchase and issue price, burdening the exchequer with a large subsidy bill.

---
[12] According to the Gulati and Saini (2013), there is about 40 percent grain leakage in the PDS.

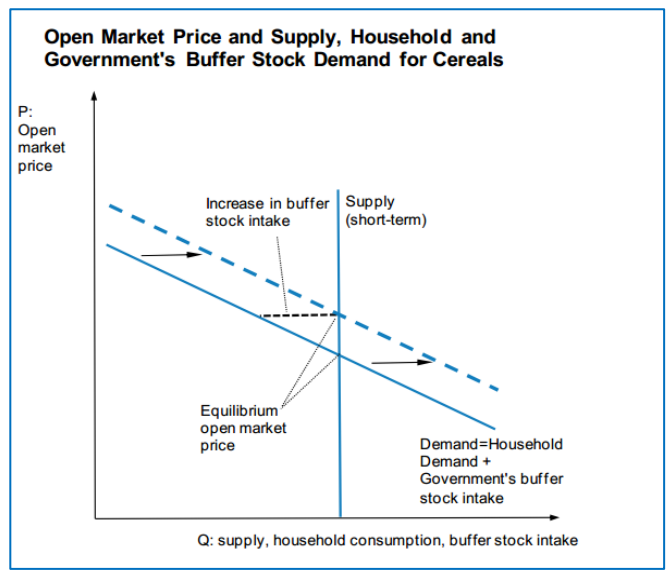

**Figure 4.** Impact of increased buffer stock intake on market prices (*Source*: Anand et al. 2016)

## 2.3 Farming and Agricultural Markets in India

As India's population and urban land grow in sizes, net sown area under crops has declined after the economic liberalisation beginning from FY91 (see Fig. 5). The average size of cultivated plots has shrunk by nearly half from 2.28 hectares in 1970-71 to 1.16 hectares in 2010-11 (Fig. 5) owing to increase in number of land holdings. Farming on such small areas is inefficient which is further aggravated by certain state laws which limit the area of agricultural land an individual can hold. Due to history of exploitation of peasants by landlords during medieval period, state laws favouring strong tenancy rights make leasing agricultural land very difficult in India. As compared to other major agricultural producers around the world, agricultural yield per unit area is fairly low in India (Fig. 6). These factors along with rise of manufacturing and service sector have caused the share of agricultural labourers in total workforce to drop significantly over the years (see Table 1). Additionally, the long-term shifts like increase in share of land use for export-oriented commercial crops since the liberalisation of economy in mid-1990s has adversely affected the growth of food output (Sonna et al. 2014).

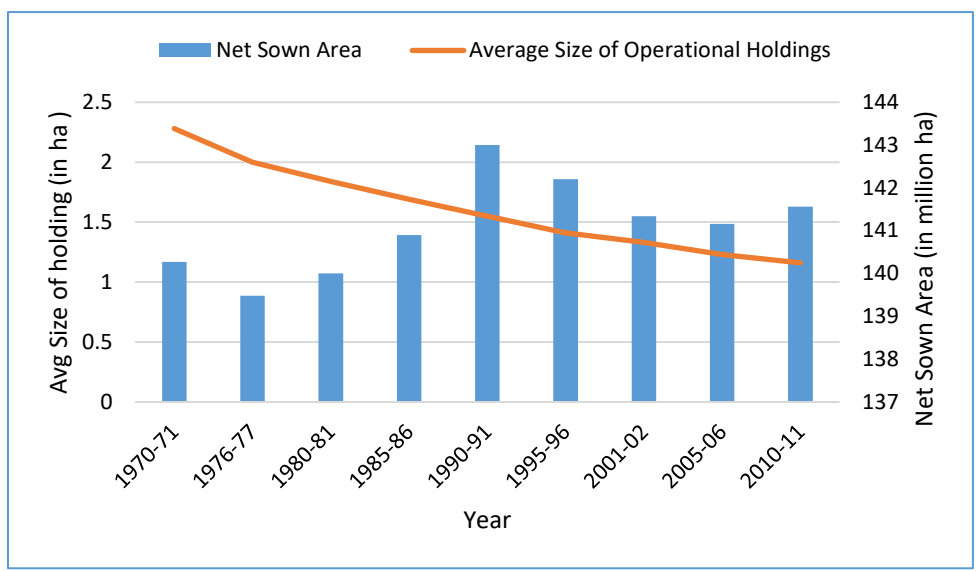

**Figure 5.** *Source*: Ministry of Agriculture and Farmers Welfare, GOI

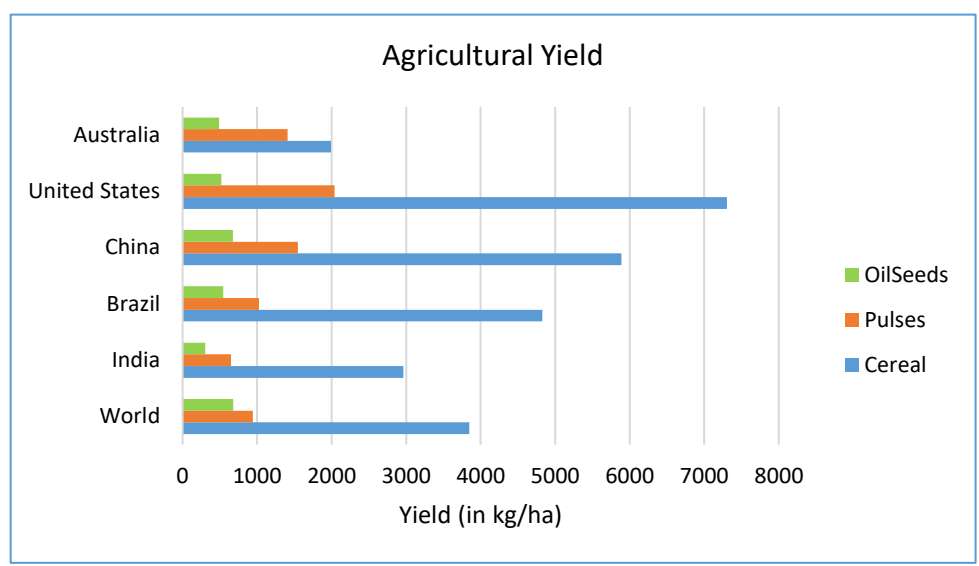

**Figure 6.** *Source*: Statistics Division, FAO

**Table 1:** Sector-wise Share in Employment (Percent)

| *Sectors* | 1999-2000 | 2004-05 | 2009-10 | 2011-12 |
|---|---|---|---|---|
| **Agriculture** | 60 | 57 | 53 | 49 |
| **Secondary sector excluding construction*** | 12 | 13 | 12 | 14 |
| **Construction** | 4 | 6 | 10 | 11 |
| **Services** | 24 | 25 | 25 | 27 |
| **Total** | 100 | 100 | 100 | 100 |

*Includes manufacturing, mining and quarrying, electricity and water supply
*Source:* Rounds of NSSO employment survey

The supply-chain facilitating movement of produce from farms to fork is severely distorted with presence of multiple intermediaries, poor logistics, price information asymmetry and addition of brokering costs at each link. The lack of adequate transportation and storage facilities leads to significant losses in the supply-chain. The monetary value of post-harvest loss incurred was found out to be in excess of INR 900 billion for 2012-13 with perishable items (fruits, vegetables and livestock produce) making up more than half of these economic losses (Jha et al. 2015). The losses in cereals and oilseeds is mainly concentrated in farm operations with relatively small losses in storage channels (see Table 2). However, the losses incurred in storage networks become significantly comparable to that of farm operations in case of fruits and vegetables.

**Table 2** Harvest and post-harvest losses at national level (*Source:* Jha et al. 2015)

| Crop type | Farm operations | Storage Channels | Total loss |
|---|---|---|---|
| *Cereals* | 4.37 % | 0.89 % | 5.26 % |
| *Pulses* | 5.79 % | 1.32 % | 7.11 % |
| *Oilseeds* | 4.73 % | 0.79 % | 5.52 % |
| *Fruits* | 7.44 % | 2.30 % | 9.74 % |
| *Vegetables* | 6.58 % | 1.98 % | 8.56 % |

The Agricultural markets are fractured in themselves led by state marketing boards known as Agricultural Produce Market Committee (APMC) which restricts farmers' trade options only to traders or commission agents licensed to operate in the area under a particular APMC. As per the Indian Constitution, agricultural marketing is a state (provincial) subject. The intra-state trading falls under the jurisdiction of state government while the inter-state trading comes under Central government. As a result, agricultural markets are prevailed and administered mostly under the several State APMC Acts. Until recently, a trader in northern state of Punjab was not allowed to bid for coconuts in southern state of Kerala. This gives opportunity to *arhatiyas* or local commissioning agents, the infamous intermediaries who add little or no value to the supply chain and are known for exploiting farmers and charging hefty commissions on sales, in some cases even up to 14 percent as opposed to the international norm of 0.5 percent on such sales. In addition to low profitability, the uncertain trade policies practiced by central and state governments discussed in the next section, further discourage farmers to invest or specialize in advanced farming techniques.

## 2.4 Agricultural Trade Policies

Food inflation is a political scandal in India as it distorts the consumption basket of a common voter. Amidst political pressure, governments respond to surge in food prices by imposing export bans, sometimes even at state levels[13]. India cautiously regulates its agricultural trade by frequently imposing import-export bans and change in duties to protect the interests of domestic farmers, relevance of MSPs as floor prices and, of course to shield the domestic markets from global price bouts. In the wake of 2007-08 global food price crisis, India adopted a very restrictive trade policy including ban on exports of wheat and rice which continued till 2011. In case of pulses, for which domestic production is not sufficient, India currently prohibits export of most of the pulses complementing this with zero import duty.

However, in events of sporadic shortages arising out of droughts etc. the delays in announcement of policy changes to meet domestic demand with increased imports often results in government agencies importing at much higher prices as the global market has already factored in the existing shortage in India. The primary reason behind this recurrent policy failure is lack of an institutional mechanism for forecasting global and domestic food supply and prices based on which timely warnings could be issued to relevant agencies. Consequently, central and state governments fail to coordinate and reach a timely solution facilitating quick imports before build-up of domestic shortage. There have been instances of temporal trade imbalance (see Chand 2010) in the past when a certain commodity was exported during the times of surplus supply at low prices and then imported back in subsequent time periods at higher prices to meet domestic shortage, resulting in huge losses to the exchequer.

---

[13] For instance, in 2014, state government of West Bengal imposed a ban on shipping potatoes to other Indian states in response to escalated prices.

India is a net exporter of cereals, but a major importer of oilseeds and pulses, in fact being the largest consumer of pulses India has become the largest producer and importer of pulses as well.

## 2.5 Food Inflation: Timeline

Elevated levels of persistent inflation have posed a serious macroeconomic challenge to India's growth prospects. The average year-on-year aggregate inflation (7.4%) and food inflation (7.2%) were at comparable levels during FY91 to FY06, but post 2007 financial crisis food inflation has always exceeded aggregate retail inflation by more than two percentage points, barring two years of FY11-12 in which crude oil prices rose sharply taking fuel component of total inflation with itself (see Fig. 1). Food prices grew at sustainable rates during 1980s and 1990s owing to success of *'Green Revolution'* - a combination of demand and supply side interventions focussed on agricultural infrastructure development, farm input subsidies and technological investments which aided in stabilizing the growth of agricultural output at par with demand. However, as growth in agricultural output slowed after 1990s, Indian government had to tap into buffer stocks to meet increasing food demand which kept the food prices in check during early 2000s. This led to depletion of stocks being held in central pool which was further accelerated with Indian government's response to shield domestic markets from surge in international food prices during 2007-08 (Fig. 3, Fig. 14). Eventually, the stocks in central pool fell below the accepted norms which prompted the food authorities to ramp the procurement causing shortage in the domestic markets. This shortage was further aggravated by low agricultural output from drought in 2009 causing food inflation to touch double digits during FY09-10. The effect of decline in global food prices in FY10 was not transmitted into the domestic market as the domestic food prices grew by more than 15% in that year primarily due to excessive stock hoarding in the wake of 2007-08 global food crisis. Even though FY10-11 had a good monsoon, the food inflation didn't ease much as both the core and aggregate inflation picked up in these years.

With a moderate gain in FY12, food prices started rising again as the household inflation expectations remained at persistently high levels during FY10-14 (Fig. 7), a period which witnessed surge in crude oil prices and political upheavals with state and general elections being held in 2013-14.

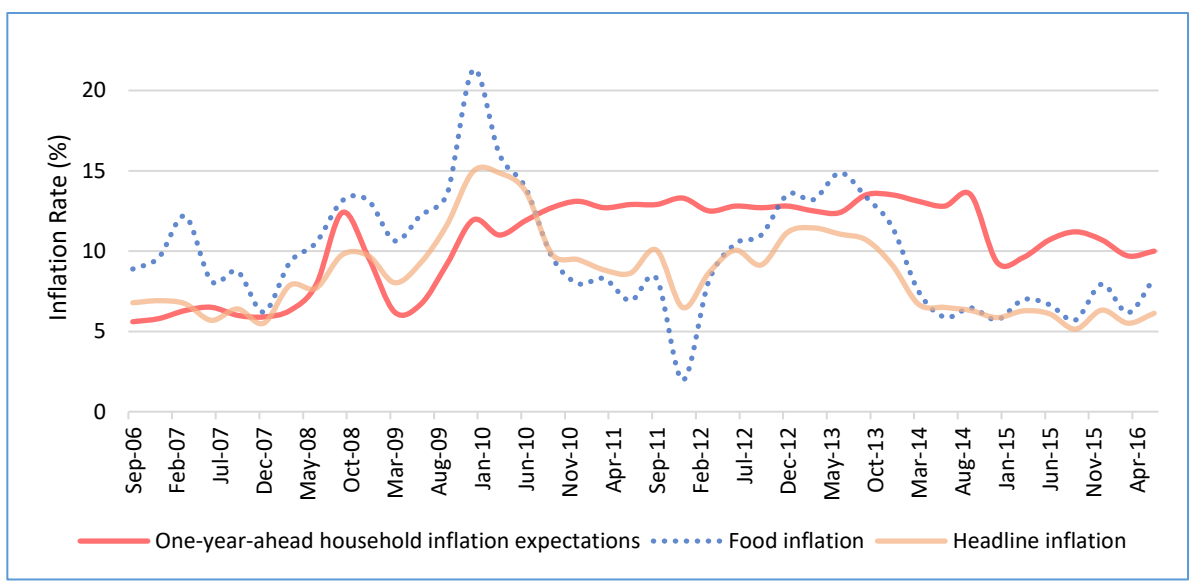

**Figure 7.** Household inflation expectations (*Source*: RBI)

Various government interventions including significant hikes in MSP, pre-election policy announcements involving food and fertiliser subsidies and other populist measures caused ballooning of central and state fiscal deficits. These measures not only caused inflation in the immediate years, but also prolonged the inflationary pressures in the economy resulting into double digit food inflation during FY13-14. With the creation of an inflationary spiral of elevated inflation expectations[14] and food inflation levels transmitting into core inflation and wages[15], the aggregate retail inflation remained at uncomfortably high levels during 2010-14 averaging

---

[14] According to RBI (2014), a one percent increment in food inflation is followed by an immediate rise in one-year-ahead household inflation expectations by half percentage points, the effect of which persists for eight quarters.
[15] The influence of Indian food inflation on price expectations and wage setting create large second-round effects on core inflation (Anand et al. 2014)

more than ten percent. Within the food items, the highest inflation was observed among pulses with a three-fold rise in prices, whereas the price of overall food basket has doubled in the last ten years.

## 2.6 Dynamics: Supply-Demand Mismatch

Driven by strong economic growth in late 2000s, per-capita real private consumption expenditure in India grew annually at an average rate of 6.7 percent during FY06-FY12. Moreover, the growth in per-capita private consumption and disposable personal income remained significantly unaffected from the repercussions of Financial crisis in 2007-08 (see Fig 8). However, with economic slowdown in FY13-14 demand-side pressures on food prices eased off as growth in consumption fell below 5 percent.

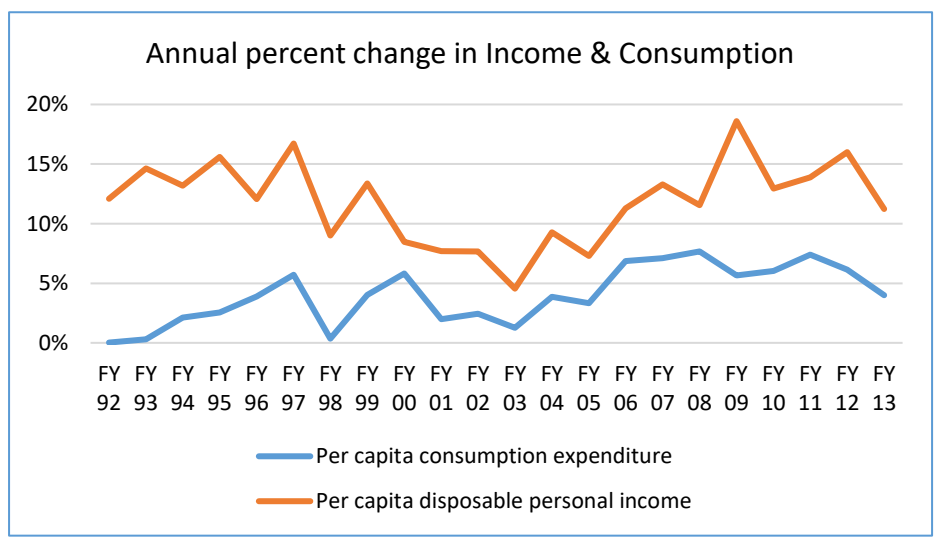

**Figure 8.** Base 2004-05 (*Source*: National Account Statistics, MOSPI)

The domestic demand for food has been rising continuously with an ever-increasing population. The domestic production has failed to keep up with this persistent rise in demand and has almost stagnated in the last five years. Domestic agricultural output grew annually at an average rate of 0.9 percent during FY01 – FY10 (Fig. 9), whereas in the same period, the population recorded an annual average growth rate of more than 1.5 percent. This mismatch in demand and supply is fairly severe in the case of pulses for which consumption has risen significantly as compared to domestic production. India has not able to meet its demand for pulses even after imports. As a result, the per capita net availability of pulses, after incorporating imports and exports, has fallen significantly from 25.2 kg/year in 1961-62 to 13.1 kg/year during 2000-2014, whereas cereal availability has increased in the same timeframe (see Fig. 10).

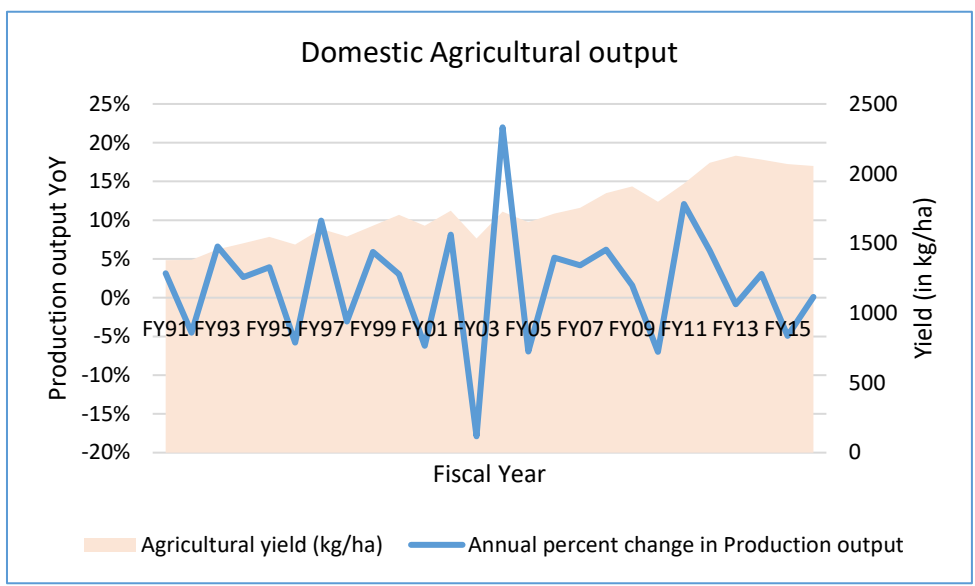

**Figure 9.** *Source*: DBIE, RBI; and author calculations

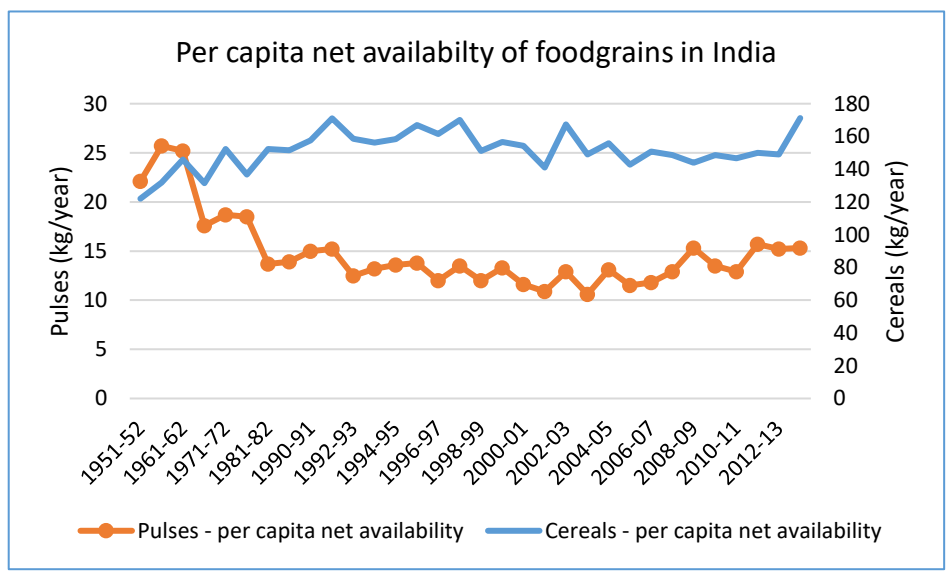

**Figure 10.** *Source*: Economic Survey 2015-16

When dealing with traded goods, it is commonly assumed that supply is perfectly elastic in prices, with demand getting adjusted to clear markets. However, as previous studies (Kumar et al. 2010) on agricultural commodities have shown that when compared to own price elasticities of demand, supply has lower own price elasticities in Indian context. The implication of this observation is more prominent in the near-term dynamics, where the movement in prices caused due to shift in demand is supposed to remain unaltered with supply response. This apparent agricultural *supply inelasticity*[16] is known to create inflationary pressures on food prices in India as noted by Reddy (2013) along with two other supply-side reasons - one of them is *supply bottleneck* which is mainly endogenous to the system, arising out of lack of adequate investment in supply logistics and supporting mechanisms; and the other being *supply shock* which is a transient endogenous factor caused primarily due to deficient monsoon, floods or other anomalies in climate.

## 2.7 Pulses: Catching fire

The last few years have witnessed frequent episodes of absurdly high inflation in pulses, with prices getting tripled in the last decade. In FY16 alone, prices of pulses increased by more than 45 percent (Fig. 11). Over the years, the supply of pulses has failed to keep up with increasing demand arising primarily out of higher rural household incomes and shift in dietary patterns. In recent times, consecutive monsoon shocks of 2014 and 2015 have further aggravated the situation and with international pulse prices at elevated levels and weak rupee, imports failed to ease inflationary pressures.

Over the long term, this shortfall in production occurred mainly due to decline in farm area under pulse cultivation with cultivators opting for high-yielding crops, such as rice and wheat, which offer higher procurement prices. This has eventually led to lower yields in pulse cultivation as it got pushed to poorly-irrigated smaller farms with low quality soils. These farms with inadequate access to irrigation are dependent on rainfall which increases the risk of crop failure. As unshelled pulses have a low shelf life, risk of post-harvest losses due to subpar storage infrastructure, limited access to milling facilities and lack of government assurance for procurement further discourages farmers to cultivate pulses on their land. Unlike rice and wheat, MSPs of pulses are much lower than open market prices. Every year, procurement prices are announced by government before the start of sowing season of pulses, just like rice and wheat, but without any procurement being done post-harvest. The lack of buffer stock for pulses certainly limits the ability of government to mitigate any adverse supply shock and control price spikes. Consequently, this has led to larger dependence on international markets which is thin and volatile in case of pulses. India had to import 5.79 MT of pulses in FY16 to meet the shortfall. The domestic production for pulses in FY16 stood around 17 MT which is significantly short of the estimated demand of nearly 24 MT. The issue of pulses is certainly an important one, as India is now the largest producer, consumer and importer of pulses globally. Unlike oilseeds which has a robust international market and involves large public and private import houses, the global market of pulses is feeble causing absence of structured import houses for pulses in India. This restricts the ability of Indian food management authorities to ease persistent inflation in pulses with international trade.

---

[16] In an inelastic supply system, only price but not supply adjusts to equilibrate demand.

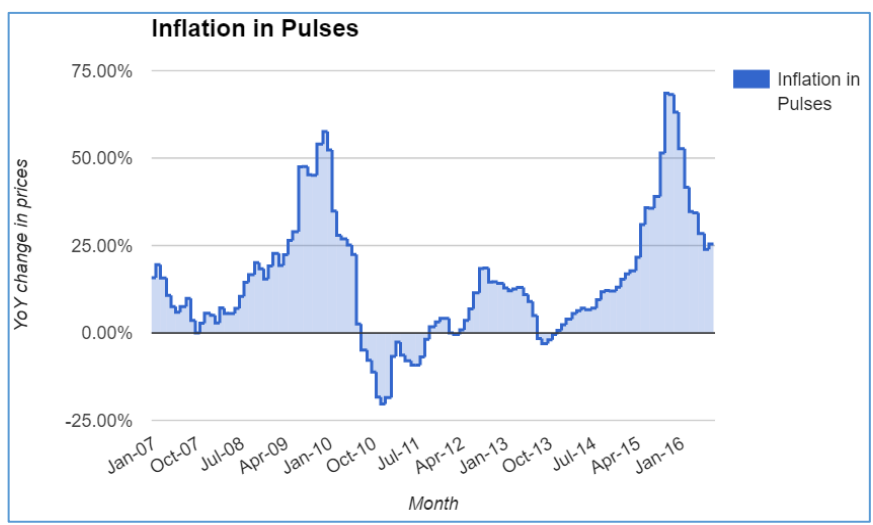

**Figure 11.** Retail inflation in pulses based on CPI-IW; *Source*: CSO

## 2.8 Measures of Inflation in India

Currently, four measures of retail inflation (CPI) exist in India along with a wholesale price index (WPI). However, data collection methodology is more robust for CPI compared to WPI, which lacks pan-India data collection centres and fixed periodicity of surveys. Moreover, CPI weightages are more closely related to household expenditure and reflects the true cost of maintaining a standard of living as opposed to WPI (see Table 3). Mathematically, CPI is better able to capture the movement in prices arising out of demand side factors as well as future inflation expectations. Adhering to recommendations made in RBI (2014) by Expert Committee to Revise and Strengthen the Monetary Policy Framework, RBI has started monitoring CPI closely while framing monetary policy and adjusting interest rates. This makes CPI a preferable measure for studying inflation in Indian scenario.

However, accessing distant past series of all-India CPI poses a difficulty as the combined CPI which encompasses all segments of population was launched recently in January 2011 with 2010 as base year. Prior to advent of all-India CPI, four different price indices were estimated catering to specific segments of Indian population; namely (i) CPI-IL for industrial labourers, (ii) CPI-AL for agricultural labourers, (iii) CPI-RL for rural labourers, and (iv) CPI-UNME[17] for urban non-manual employees. Among these, CPI-IL has a wider geographic coverage and latest base year than the other two price indices. Moreover, CPI-IL is used as a proxy of cost of living index in the organised sector and had been a broad-based inflation indicator for the country as a whole before the introduction of all-India CPI. RBI (2014) observed CPI-IL and all-India CPI showing similar inflation trends making it the most suitable measure of inflation for econometric analysis and the same has been employed in this study.

**Table 3.** Item-wise weights

| Group/Sub-group | CPI-IW (Base 2001) | NSSO 2011-12 household survey | | WPI (Base 2004-05) |
|---|---|---|---|---|
| | | Rural | Urban | |
| Cereals and Products | 13.48 | 12.1 | 7.4 | 6.28 |
| Pulses and Products | 2.91 | 3.3 | 2.2 | 0.72 |
| Milk and Products | 7.31 | 9.1 | 7.8 | 3.24 |
| Edible Oil | 3.23 | 3.8 | 2.7 | 3.04 |
| Egg, Fish and Meat | 3.97 | 3.6 | 2.8 | 2.41 |
| Other Food | 15.3 | 16.7 | 15.7 | 13.4 |
| **Food Total** | **46.2** | **48.6** | **38.6** | **29.09** |
| **Non-food Total** | **53.8** | **51.4** | **61.4** | **70.91** |

---

[17] CPI-UNME was discontinued permanently in 2010.

# 3. Factors driving Food Inflation in India

## 3.1 Monsoon dependency

The very nature of Indian economy puts in perspective significance of rains in making a good crop year, especially southwest monsoon lasting from July to September. More than half[18] of India's net sown area, land which is cropped at least once a year, still remains unirrigated and relies on water that rains down from the clouds, mostly in the monsoon months. Over the time, deviations in geographical distribution and rainfall observed during southwest monsoon have significantly affected the agricultural output and hence food prices (Mohanty, 2014). Analysing the historical data (between 1956 and 2010), Mohanty (2010) argues that high food prices caused by droughts were responsible for more than three-fourths of the instances of double digit inflation in India. A deficient monsoon builds inflation expectations and adversely affects the production output. Empirical evidence based on data from past 25 years suggests a positively weak correlation of 0.3 between food inflation and monsoon deviation from long term mean in the same year, which is counterintuitive in itself (ideally, it should come negative!). However, on segregating the drought years[19], a significantly negative correlation of -0.8 is found out between monsoon deviation and food inflation in the following year (Fig. 12).

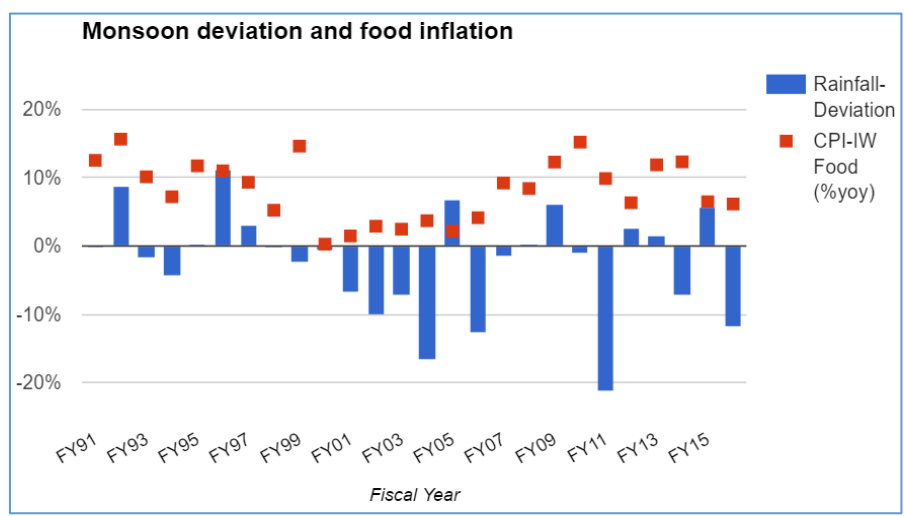

**Figure 12.** Food inflation based on CPI-IW (base year 2001) (*Source*: DBIE, RBI)

---

[18] *Source:* Ministry of Agriculture & Farmers Welfare, Government of India.
[19] For the purpose of this study, year in which monsoon was deficient by more than 10 percent from long term average was considered as a drought year.

## 3.2 Minimum Support Prices

The role of MSPs in guiding food inflation is fairly large as the crops covered under MSP constitute more than a third of all-India food consumption basket[20]. The MSP as a concept is intended to be a floor for market prices but during years with substantial hikes, it eventually ends up setting market prices directly, which is generally followed by rise in prices of key agricultural crops (Rajan 2014; Mishra and Roy 2011). With the abolishment of practice of announcing separate procurement prices in mid 1970s, MSPs have become minimum prices at which FCI stands ready to buy/procure whatever quantities farmers have to offer. This certainly has aided in excessive building of buffer stocks in central pool when hikes have been significant.

The use of incentive schemes like MSP in boosting agricultural production may be limited as Rajan (2014) argues that, "the gains from MSP increases have not accrued to the farm sector in full measure on account of rising costs of inputs". This is evident from the trend in internal Terms of Trade[21] (ToT) of agricultural commodities which has flat-lined in the recent times. Rajan (2014) compares the approach of increasing production through hikes in MSP to "*a dog chasing its tail*" - it can never catch it, as hikes in MSPs also drives input costs upwards. However, if higher MSPs lead to higher supply of rice and wheat, which are the primary commodities procured at MSP, then this might also result in a suboptimal mix of production, distorted towards rice and wheat, with a reduced supply of other food commodities causing non-cereal food inflation.

As shown in Fig. 13, an MSP-induced increment in production is necessitated by a subsidy to a consumer to ensure post-harvest market clearing of increased cereal supply assuming no buffer stock build-up. Thus, given the increase in fiscal burden arising out of food subsidies, the increase in MSPs in combination with open ended procurement and PDS may elevate overall inflationary pressures in the economy even in a scenario of increased production and declining food inflation.

---
[20] *Source*: All India Weights of different Sub-groups within Consumer Food Price Index - 2012 series
[21] Terms of Trade is defined as ratio of changes in input cost over the changes in the output price of agricultural commodities. A constant ToT over time implies towards a stagnant profitability, thereby, reducing incentive for investment in agriculture.

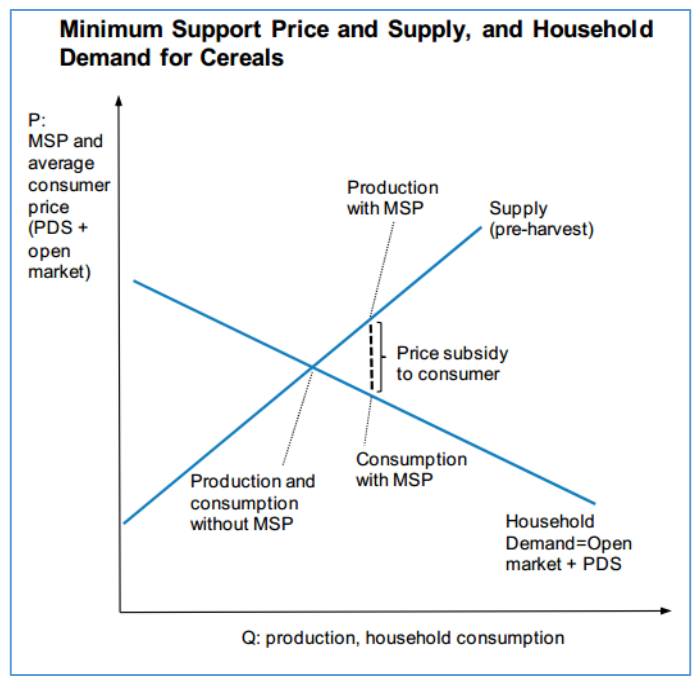

**Figure 13.** Impact of MSP hike on demand-supply dynamics (*Source*: Anand et al. 2016)

## 3.3 International Prices and Trade Policy

After 1991, when economic reforms towards opening the economy were implemented, agricultural markets in India have been progressively integrating with global markets. Consequently, a shift in international food prices exert direct and indirect influence on domestic markets through trade as well as through policy adjustments. Historically, the rising trend in global food prices have been slowly transmitted into the domestic market (see Fig. 14) owing to India's cautious trade policy. A massive spike in food commodity prices was observed in 2007-08 which caused cereal prices to almost triple from their early 2000 levels droving 44 million people into poverty around the globe (World Bank, 2011). Many structural reasons have been identified causing this politically destructive crisis[22] ranging from spike in energy prices[23] to increasing biofuel subsidies, and even the role of global fertiliser cartels[24].The causes may be many but the effect, however left a grave imprint on food management policy in China and India for the subsequent years which recorded unprecedented grain procurement and hoarding.

---

[22] The resulting discontent from food shortages played a significant role in causing riots and protests during 2011 *Arab Spring* that toppled governments first in Tunisia, followed by the Tahrir Square uprising overthrowing Egypt's government and civil wars in many other Arab nations.

Overall, there appears to be low degree of correlation between global and domestic food prices, however, on observing closely it is apparent that a higher correlation exists during periods of low global food inflation, indicating a weak pass through during periods of high global food inflation. This makes food prices in India less volatile than their international counterparts, the trends, however, ultimately converges over the long-run implicating at impermanent success of India's agri-trade policy designed to shield domestic markets from global price spikes. Interestingly the correlation between trends have weakened in the recent years as the inflation in international food prices have eased, even entering the negative territory in case of some food commodities contrary to Indian food prices which are on a persistent rise. This weak correlation could possibly be due to large fluctuations in USD-INR exchange rate post-2008 crisis. On comparing the domestic food index with international food price index denominated in INR, the two indices seem to follow a similar path (Fig. 14).

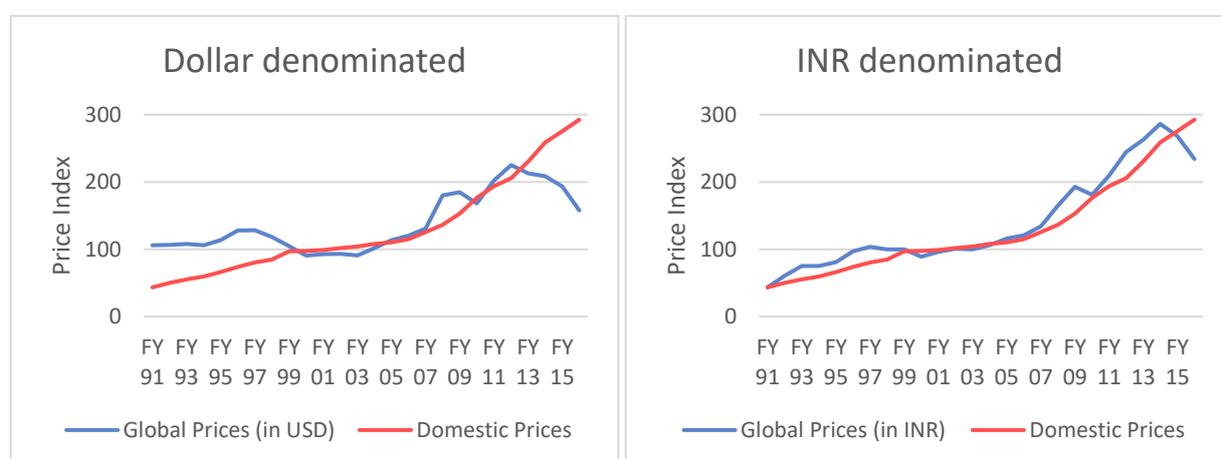

**Figure 14.** Movement of domestic prices with global food prices (*Source*: FAO; DBIE, RBI)

---

[23] Agriculture consumes energy both directly, through usage of diesel in operating farm machinery, and indirectly, through natural gas providing nitrogen for fertiliser production.

[24] In recent studies, Gnutzmann and Spiewanowski (2014; 2016) exposed the role of fertilizer cartels in elevating food prices. They estimated that the pure fertilizer cartel effect explained more than 60% of rise in food prices during the crisis and food prices would have been around 35% lower at the peak of crisis, has the fertilizer prices were set competitively in the absence of a cartel.

## 3.4 Fiscal Policies

In general, prices of agricultural commodities tend to be more responsive to macroeconomic shocks as compared to several manufactured goods whose prices are bound by long-term contracts (Thompson, 1998). Particularly, in a developing economy, where demand for food is relatively inelastic and short-run price elasticities of agricultural supply is still lower, any shift in demand arising out of macroeconomic policy change, such as fiscal or monetary stimulus, could distort food prices significantly. A similar situation arose post 2007-08 financial crisis, prior to which India had successfully brought down its Fiscal Deficit from almost 10 percent of GDP in FY02 to 4 percent in FY08 (Fig. 15) following the impressive fiscal consolidation with the passage of 2003 FRBM Act. The revenue deficit became almost null while the primary deficit achieved a marginal surplus in FY08 (Fig. 16). But as the fears of severe unemployment and recession grew in the wake of 2007-08 financial crisis, G7+5 countries as well as IMF favoured a fiscal stimulus up to 2 percent of GDP as a pathway out of the feared recession.

There was a striking difference between the expansionary fiscal policies adopted in India and other countries, for instance in China, nearly 40 percent of the fiscal stimulus was directed toward investment in infrastructure development whereas, the Indian fiscal package was more focussed on stimulating demand by granting direct subsidies (ranging from food, fertiliser, energy subsidies to even flat waivers in agricultural debt[25]), expansionary income support schemes like MGNERGS[26] and generous pay hikes announced in Sixth Pay Commission for state servants. These welfare and employment guarantee schemes imparted substantial amounts of liquidity and purchasing power, particularly to rural households, boosting demand for food items (Rakshit 2011) but with several supply bottlenecks in place, particularly, stagnant productivity and subpar infrastructure, the situation soon got transformed into demand-pull inflation. The withdrawal of these politically sensitive outright doles still poses a challenge in winding down the deficit financing without an abrupt shock to an already fragile growth.

---

[25] Just before 2009 general elections, then ruling United Progressive Alliance (UPA) government waived the repayment of loans made to small and mid-sized farmers which, some commentators believed, helped the coalition get re-elected (Rajan 2011)

[26] Mahatma Gandhi National Rural Employment Guarantee Scheme

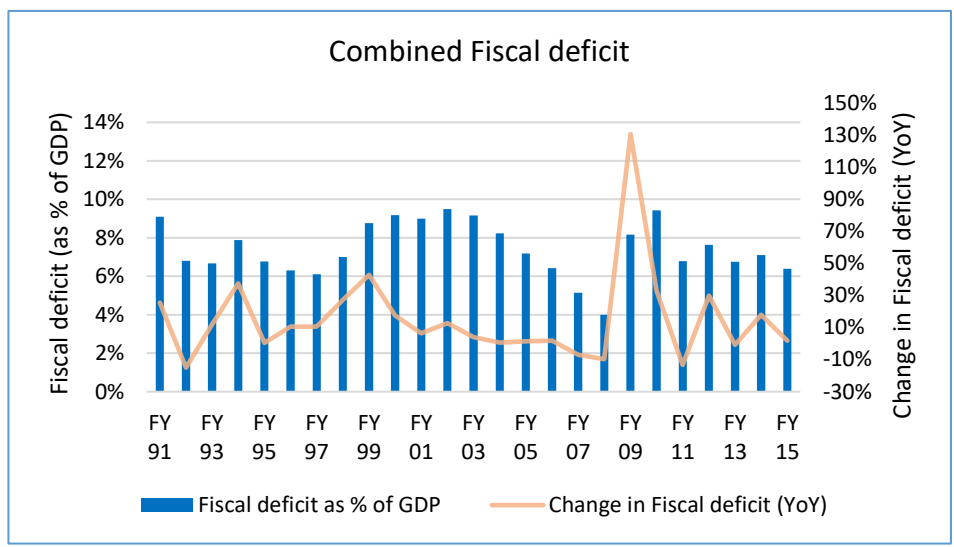

**Figure 15.** Study of combined central and state governments fiscal deficit (*Source*: MOSPI)

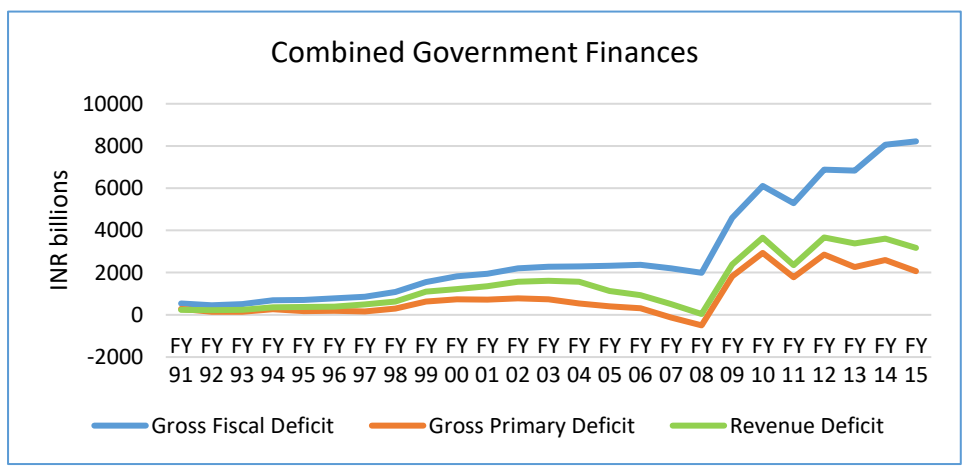

**Figure 16.** Study of combined finances of central and state governments (*Source*: MOSPI)

## 3.5 Rising Input costs

Rise in farm wages inflates the cost of production and is believed to cause a wage spiral in the economy by raising the benchmark *'reservation wage'* - the lowest rate that workers are prepared to accept for jobs across sectors which could possibly increase demand for food as well (Gulati and Saini 2013). Change in farm wages disturbs the food price equilibrium in an unbalanced way as low rural wages suppress demand only to a certain extent owing to the relatively low income elasticity for food expenditure of rural households (which constitutes 70 percent of Indian population). On the other hand, rise in rural wages not only boosts demand but also rejigs the food basket towards higher value and protein-rich items, thereby disturbing both the demand (strong rural demand) and supply (high labour costs) side of price equilibrium.

Farm wages have witnessed a sharp rise in the last decade, with nominal wages growing annually at 15 percent. Nevertheless, on deflating these wages with CPI-AL, real wages still grew at a rate of 5.3 percent per annum. Several reasons are responsible for such substantial rise in farm wages including strong private investment spurred by an increase in commercial credit flowing to the agricultural sector, scarcity of agricultural labourers caused by a shift in employment pattern with labour force moving from agriculture to non-agriculture sectors, particularly construction (see Table 1) and implementation of MGNREGS.

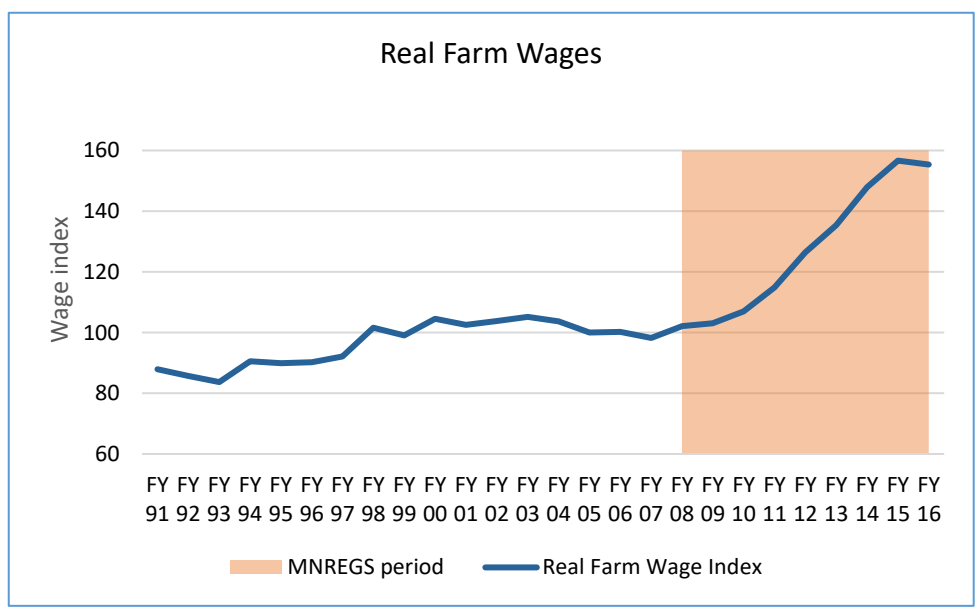

**Figure 17.** Growth in real farm wages (*Source*: Indian Labour Bureau)

The role of MGNREGS in accelerating farm wage inflation is, however, still debated in the literature. The latter phase of high growth in farm wages certainly coincided with the implementation of MGNREGS in FY08 (see Fig. 17). Guaranteed wages under MGNREGS may have strengthened the bargaining power of unskilled labourers in rural areas, but past empirical studies show that only a small fraction of increase in rural wages could be attributed to MGNREGS and that small effect, if any, is ebbing in recent years (Sonna et al. 2014).

However, the rising trend in real MGNREGS wages suggests that its effect on rural wage inflation will not disappear entirely; a recent study on the southern state of Kerala (Dhanya 2016), revealed that implementation of MGNREGS caused a rise in wages for certain rural works carried out by female labour force and was accompanied with increased food consumption expenditure.

Apart from labour costs, CACP takes other factors of agricultural production into account, as well, while recommending MSPs to the government. Non-labour operational costs contribute nearly half to the total cost of cultivation. The cumulative non-labour agri-input price index has recorded more than 80 percent rise during FY05-FY16.

## 3.6 Shift in Food Consumption patterns

NSSO's periodical survey on household consumption reveals that share of food in total consumption expenditure has been declining. The observed pattern is in line with Engel's law which states that as the average household income rises, share of food in total expenditure diminishes (Fig. 18) owing to relatively low income elasticity for food expenditure. Despite the declining overall share of food in total consumption expenditure, real per capita food consumption has been rising, especially in rural households (Rajan 2014). With rising incomes, dietary preference has shifted in accordance to Bennet's law towards more nutritious and high value food items away from starchy cereals (Fig. 19). This changing preference has resulted in high levels of inflation in pulses and other protein-rich items in the recent years. The share of protein-rich food items, both plant and animal, in total food consumption has risen to almost 40 percent. The prices of nutrition-rich items, including vegetables and fruits, have recorded a higher inflation than cereals owing to suppressed growth in supply of these items, causing a mismatch with rapidly growing demand (Mohanty 2011 and 2014). Evidence from the literature (Rajan 2014; Gokarn 2010 and 2011) asserts that protein and high value food items have become prominent determinants of overall food inflation in the recent times.

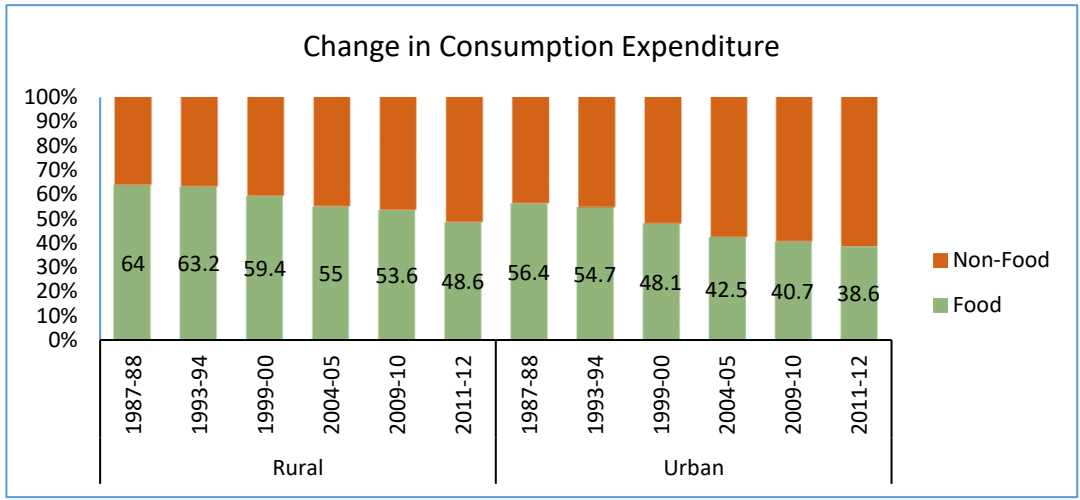

**Figure 18.** Change in consumption pattern (*Source*: Rounds of NSSO Survey)

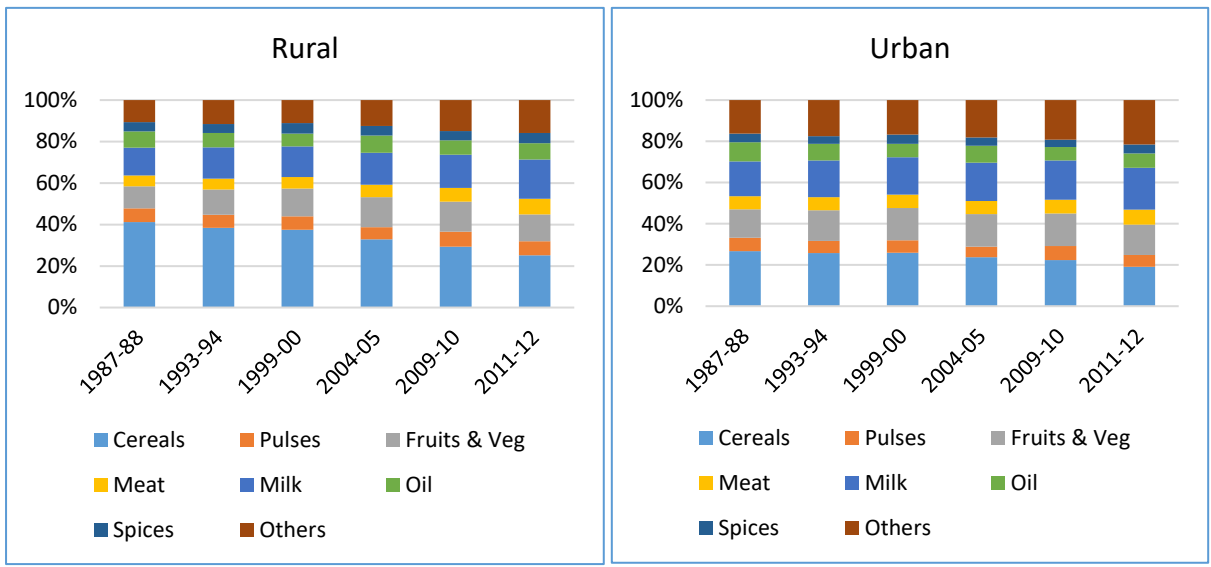

**Figure 19.** Change in dietary pattern (*Source*: Rounds of NSSO Survey)

# 4. Statistical Modelling Framework

To study the relative significance of individual factors in explaining the inflationary trend in food prices, a nonparametric regression ensemble has been developed using decision trees with gradient boosting for least squares (Friedman 2001). Unlike parametric regression, supervised machine learning techniques, such as gradient boosting, do not attempt to characterize the relationship between predictors and response with model parameters but rather produces an ensemble of weak prediction models, decision trees in this case, such that it improves the overall predictive performance of model (Hastie et al., 2009). Boosting, as a concept, provides sequential learning of the predictors, where each decision tree is dependent on the performance of prior trees and learns by fitting the residuals of preceding trees. Unlike linear regression, regression with boosted decision trees can model complex functions by accounting for non-linearity and interactions between predictor variables (Müller et al. 2013). The ability of BRTs to automatically handle missing data points saves the effort of data pre-processing. The advantage of BRTs over other predictive modelling techniques is the typical combination of high predictive accuracy and robustness against overfitting.

In the present study, the annual inflation in retail food prices (FCPI), measured by year-on year change in prices of food items included in CPI-IW (base 2001), is chosen as the response variable, whereas the predictor variables[27] include:

MonsDev     : Deviation in rainfall received during southwest monsoon in a year from its 50-year long-term mean

MSP         : YoY change in production weighted MSP index of major food crops

FAO         : YoY change in FAO price index denominated in INR

FD          : Combined fiscal deficit as a percentage of total GDP in a year

FWI         : YoY change in farm wage index

AgriInput   : YoY change in price index of agricultural inputs constructed from WPI

ProteinExp  : YoY change in ratio of private consumption expenditure on protein-rich food items to total food expenditure

The expression of variables in percentage term brings all predictor variables as well as response variable on the same measurement scale.

---

[27] Selection of variables is not a priority, as BRTs tend to ignore irrelevant predictors (Elith et al.,2008).

## 4.1 Mathematical Formulation

Gradient tree boosting employs decision trees of constant size as basis functions, $h_m(x)$, referred to as weak learners in this context, for generating additive models as the following:

$$F(x) = \sum_{m=1}^{M} \gamma_m h_m(x)$$

wherein $M$ and $\gamma_m$ represent the total number of trees employed and step length respectively. The model is constructed in a progressive stepwise manner,

$$F_m(x) = F_{m-1}(x) + \gamma_m h_m(x)$$

At each step, the next decision tree $h_m(x)$ is chosen to minimise the given loss function, $L$, for the current model $F_{m-1}(x)$ and its fit $F_{m-1}(x_i)$:

$$F_m(x) = F_{m-1}(x) + \arg\min_h \sum_{i=1}^{n} L(y_i, F_{m-1}(x_i) - h(x))$$

The initial model $F_0$ is specified by the type of loss function used. The idea behind gradient boosting is to solve this minimization problem numerically through steepest descent. The steepest descent is identified as the negative gradient of the given loss function calculated at the current model $F_{m-1}$,

$$F_m(x) = F_{m-1}(x) + \gamma_m \sum_{i=1}^{n} \nabla_F L(y_i, F_{m-1}(x_i))$$

Now, the step length $\gamma_m$ is selected by means of line search:

$$\gamma_m = \arg\min_\gamma \sum_{i=1}^{n} L\left(y_i, F_{m-1}(x_i) - \gamma \frac{\partial L(y_i, F_{m-1}(x_i))}{\partial F_{m-1}(x_i)}\right)$$

In addition to conventional BRTs, a simple regularisation strategy, as proposed by Friedman (2001), has been employed to improve the accuracy of the model which scales the contribution of each tree by a factor $\vartheta$, also known as learning rate:

$$F_m(x) = F_{m-1}(x) + \vartheta \gamma_m h_m(x)$$

Stochastic gradient boosting, which combines gradient boosting with bagging, could further improve the performance of model; wherein at each iteration a subsample of data is randomly drawn (without replacement) from the available dataset (Friedman, 1999).

## 4.2 Model parametrization

The calibration of developed BRT model is done through the following parameters,

i. *Number of trees:* The number of trees/iterations determines the model complexity. A large number of trees is recommended to exhaust the internal data structure and bring the mean squared error to statistically acceptable levels.

ii. *Learn rate / Shrinkage:* Slow learning rate improves predictive performance at the expense of increased computation time, however, smaller shrinkage values are recommended while growing a large number of trees.

iii. *Maximum nodes / splits per tree:* The number of splits determines the tree complexity and order of interactions between predictor variables. In general, a tree with *k* nodes can capture interactions of order *k*. Hastie et al. (2009) recommends four to eight splits for most cases.

iv. *Minimum number of observations in terminal nodes:* This parameter should be kept below five for small datasets

v. *Subsample fraction:* The rate of subsampling decides the proportion of learn sample to be used at each modelling iteration. Subsampling fraction close to 1 makes the model computationally intensive, however values between 0.9 to 0.95 are recommended when dealing with small datasets.

vi. *Regression Loss criterion:* Among all kinds of loss functions, the natural choice for regression is least squares owing to its overlying computational attributes. For least squares, the initial model $F_0$ is specified by the mean of the target values.

## 4.3 Data

The annual data series is considered for all the variables spanning a period of 25 years starting from FY91 to FY16. The year FY91 is chosen as a starting point for a specific reason as the first round of reforms aimed at economic liberalisation began in 1991 which reduced state monopoly and led to gradual integration of Indian domestic markets with global markets. Data used in this study has been primarily imported from various official sources, viz, Reserve Bank of India (RBI), Labour Bureau of India, Central Statistics Office (CSO), Ministry of Agriculture & Farmers Welfare, Ministry of Finance and Open Government Data (OGD) Platform India. A production weighted MSP index (base: FY05) has been created for major food crops including; cereals - rice, wheat and maize; pulses - gram and tur; oilseeds - groundnut, mustard and soybean. The annual average of global food price index (base: 2002-04) published monthly by FAO (Food and Agricultural Organisation) is converted into rupee terms using the yearly average of USD-INR exchange rate. Combined fiscal deficit of central and state governments as a percentage of total GDP is considered. A farm wage index[28] (base: FY05) has been constructed by aggregating wages earned by agricultural labourers engaged in primary farm operations, namely, ploughing, sowing, transplanting, weeding, and harvesting. The weights used in constructing non-labour agri-input price index[29] (base: FY05) have been derived from WPI 2005 series. Relative expenditure on protein-rich items is based upon weights derived from Private Final Consumption Expenditure[30] (PFCE) at constant prices (base: 2003-05), for both plant and animal protein based items, namely, pulses, oil & oilseeds, milk & milk products, and meat-egg-fish (MFE).

---

[28] The data for farm wages is available from FY91, hence YoY change could only be measured starting from FY92.
[29] The non-labour agricultural inputs include agricultural electricity, light diesel oil, high speed diesel, agricultural machinery & implements, tractors, lubricants, fertilisers, pesticides, cattle feed and fodder.
[30] PFCE item-wise breakup data is available only till FY13.

## 4.4 Results and Discussion

The BRT model has been constructed over 50,000 trees with a learn rate of 0.0001 and subsample fraction of 0.95. To capture multi-variable interactions in the given small dataset, six maximum nodes per tree are allowed and minimum number of observations in terminal nodes is kept at three. The developed model fits the actual data quite well (see Fig. 20 and Table 4) with a R-squared value of 99.1 percent and MSE (mean squared error) flatlining beyond 30,000 trees (Fig. 21).

**Table 4** BRT model error measures

| | |
|---|---|
| MSE (Mean squared error) | 0.00002 |
| MAD (Mean absolute deviation) | 0.00321 |
| R-sq (R-squared) | 0.99073 |
| ROC[31] (Area under curve) | 0.99877 |

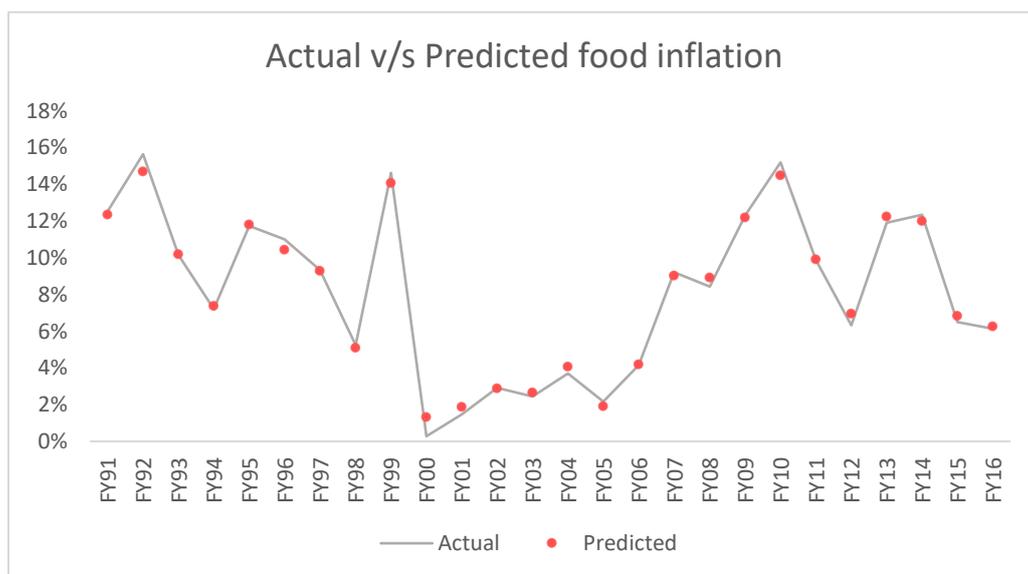

**Figure 20.** Comparison of BRT model predictions to actual data

---

[31] In machine learning, accuracy is measured by the area under the ROC (Receiver Operating Characteristics) curve, wherein the model performance is considered excellent for values between 0.9 and 1.

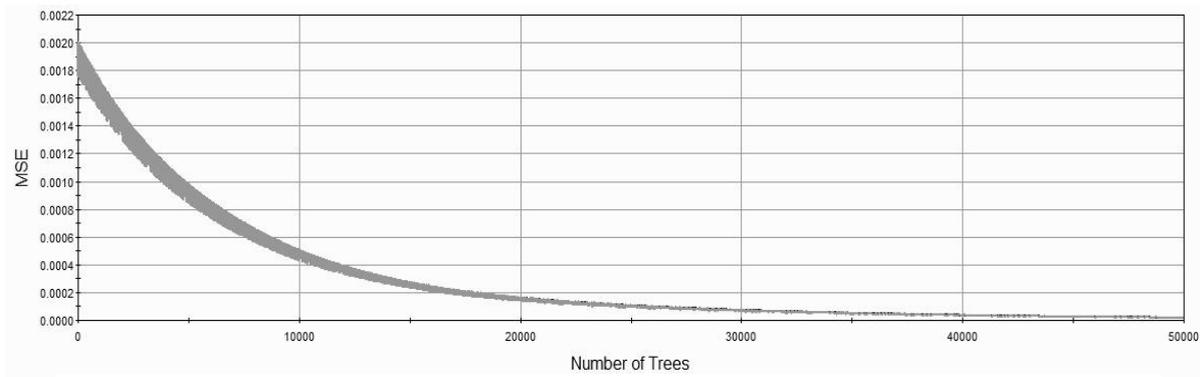

**Figure 21.** Change in MSE with growing trees

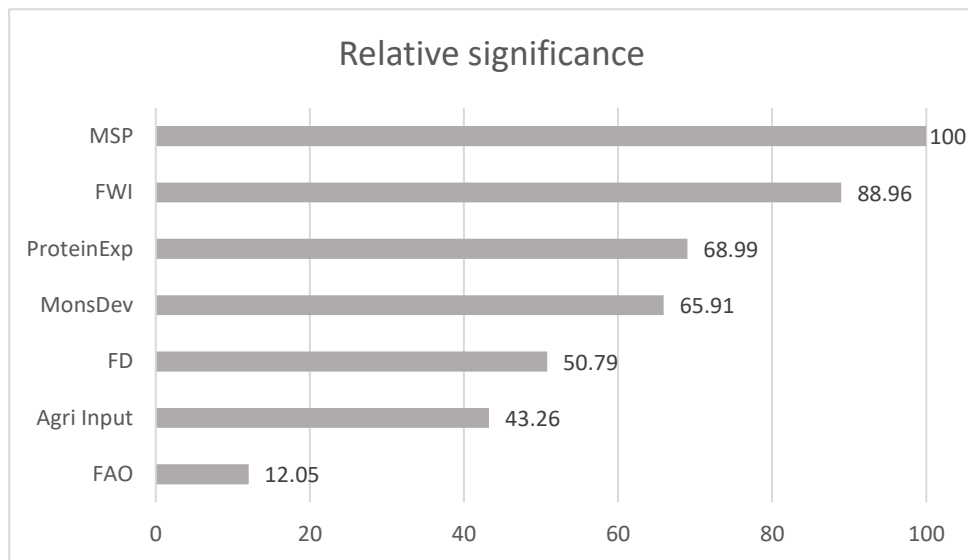

**Figure 22.** Relative significance of explanatory variables

The relative importance of the predictor variables in the model is manifested via a relative influence score of a variable, which is decided based on the number of times a variable gets selected during splitting, weighted by the squared improvements, and averaged over all possible trees (Müller et al. 2013). The results (see Fig. 22) indicate that none of the predictor variables, except probably FAO, is insignificant in explaining the inflation in food prices and contributions are fairly distributed among the variables, with MSP contributing the most to predictive performance of BRT, followed by farm wages. Minimum support prices carry most explanatory significance as they directly set the floor for market prices. Additionally, hikes in MSP not only elevates the inflation expectations but combined with PDS and employment guarantee schemes they set indirect inflationary pressures in the economy as well, by burdening the exchequer with bloated subsidy bills and inducing a wage price spiral. Farm wages have recorded a sharp rise especially from FY08 onwards and during this period the annual food inflation has averaged above 9 percent. This rise in farm wages has certainly induced cost-push inflation in food prices by raising the overall input costs of food production in India and stagnating the agricultural profitability at the same time. The low explanatory significance of FAO index may in part relate to agricultural trade policies adopted by India which prevent transmission of global food price short-term volatility into the domestic markets and allow only long-term trend in global food prices to be captured into the recommendations made by CACP for setting MSPs.

Additionally, BRT model allows visualisation of the relationships between a single predictor and the response variable via univariate partial dependence plots (Friedman, 2001; Friedman and Meulman, 2003) wherein, the predictor variable of interest is varied over its range while the remaining predictors are fixed at joint set of values sampled from the dataset to produce an instance of the response dependence curve. This process is replicated for all learn records, sampling new joint set of fixed values each time, creating a family of partial dependence curves which are then averaged and centered so as to generate the final partial dependence plot (PDP). The comparable scales of vertical axes in all plots (see Fig. 23) indicate that all the predictor variables are significant in explaining the variation in food prices, including FAO.

Contrary to recent reports published by commercial banks (Varma and Saraf 2016), which fail to find significant dependence of food inflation on monsoon deviation, the PDP generated in Fig. 23 clearly reveals food inflation reacting inversely to even small deviations in monsoon. This is particularly due to inflation expectations, which have relied heavily on the performance of southwest monsoon throughout Indian history. The employment of ordinary least squares regression methodology by Varma and Saraf (2016) to study the dependence of food inflation on monsoon without controlling for average effects of remaining predictors is perhaps what lead to discrepant findings. PDPs, however, are not perfect representation of the effects of each variable, especially when the predictors are correlated or if there exist strong interactions within the dataset (Elith et al.,2008).

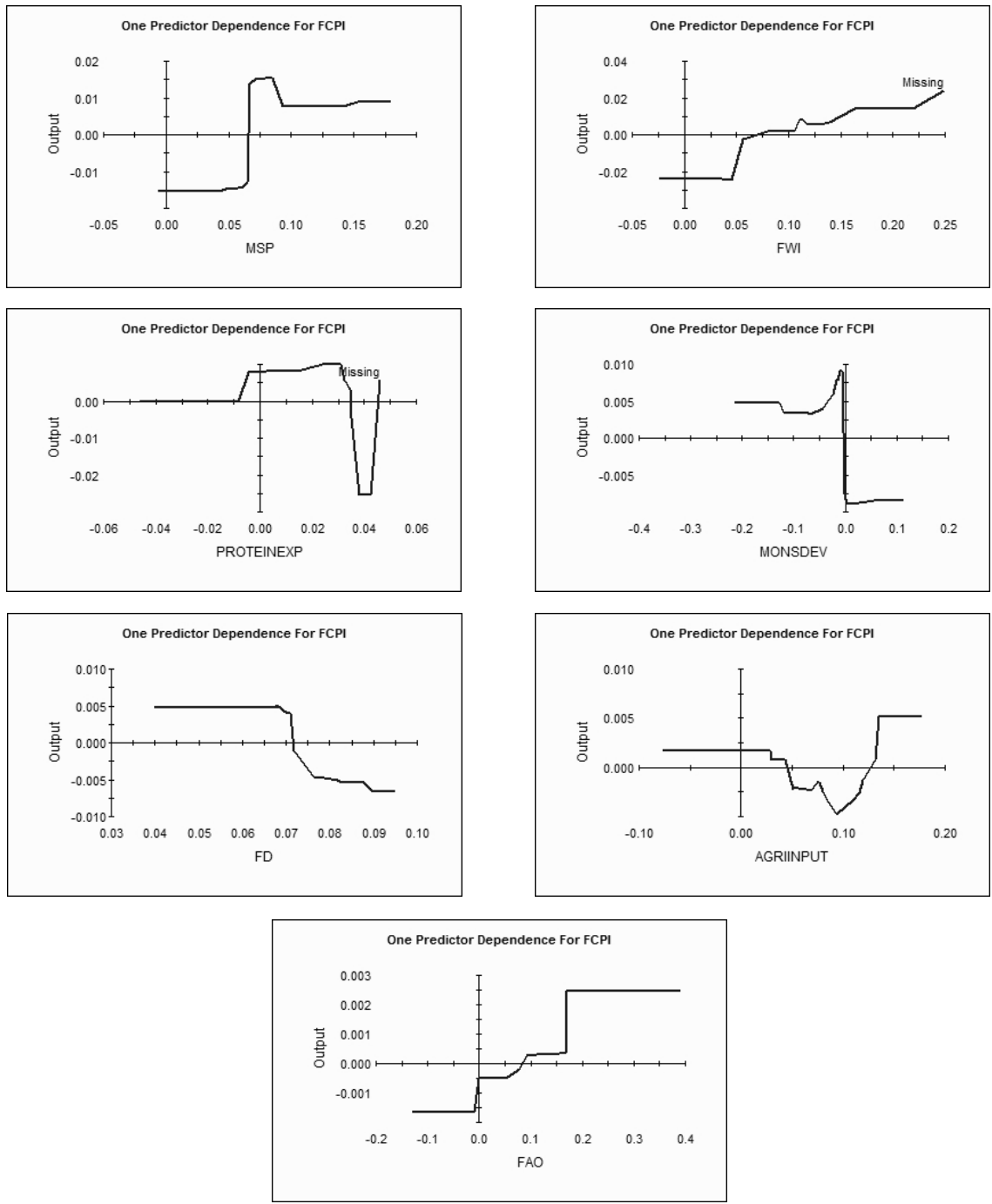

**Figure 23.** Univariate partial dependence plots for all predictors

As the interactions among predictors are allowed, the predictive performance of BRT model improves (see Fig. 24) The strength of interaction effect between a given pair of predictor variables could be empirically evaluated by gaging the difference in response surfaces of a genuine bivariate plot (allowing predictors to interact) and an additive combination (no interaction) of the two corresponding univariate plots. As Table 5 shows, four of the five most important pairwise interactions include MSP and FWI, the two most relevant predictors. The strong interaction between farm wages and protein expenditure reinforces the case of inflation driven by boost in rural consumption amidst rising farm wages.

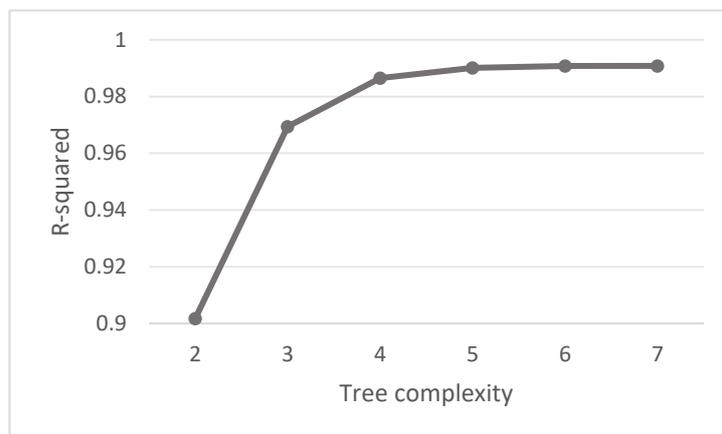

**Figure 24.** Model performance and Tree complexity

**Table 5** Pairwise Interaction Score

| Predictor I | Predictor II | Pair interaction score[32] |
|---|---|---|
| MSP | MonsDev | 12.4 |
| FWI | ProteinExp | 11.9 |
| MSP | FD | 7.0 |
| MSP | FWI | 6.3 |
| MonsDev | ProteinExp | 6.2 |

All possible pairs of variables are exhausted to arrive at the overall interaction strength of each predictor variable, which is expressed on a cumulative percent scale. MSP that captures long term trend in global food prices as well as labour and non-labour input costs emerges with the highest interaction score and FAO with the least (see Table 6). The two most significant predictors, MSP and FWI, also have the highest overall interaction strength. The interaction between these two variables could be visualized with joint partial dependence plot (Fig. 25).

---

[32] Expressed on the percent scale, pairwise score reflects the contribution of the interacting pair normalized to the total variation in the output response.

Partial dependence plots for other pairs with high interaction scores (from Table 5) are displayed in Fig. 26 (a-d).

| Table 6 Overall interaction strength of variables | | | | | | | |
|---|---|---|---|---|---|---|---|
| Predictor | MSP | FWI | MonsDev | ProteinExp | FD | AgriInput | FAO |
| Score[33] | 23.39 | 20.94 | 19.51 | 17.76 | 9.56 | 8.44 | 1.77 |

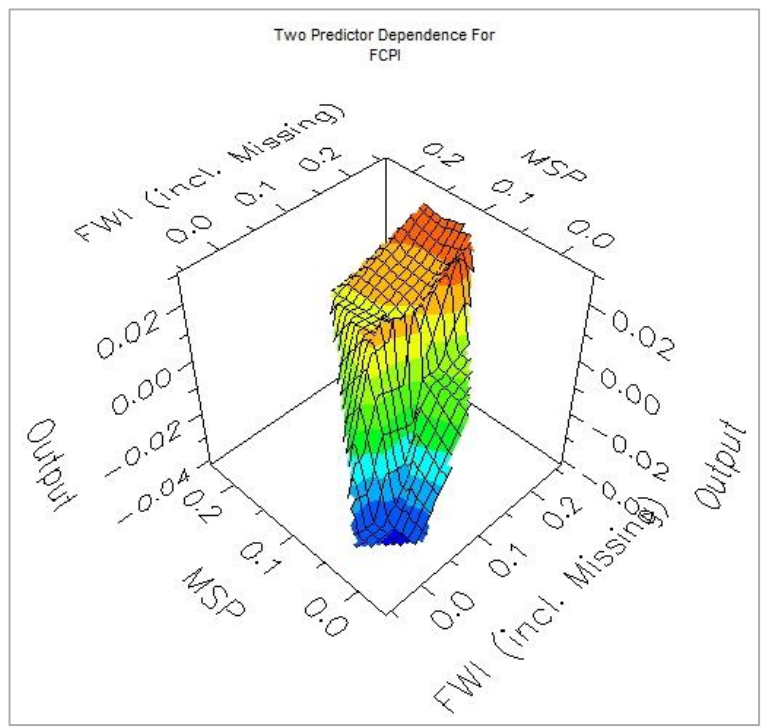

**Figure 25.** Bivariate partial dependence plot for MSP and FWI

---

[33] As an example, the overall interaction score of MSP indicates that around 23 percent of the total variation in food prices is attributable to interaction of MSP with other predictor variables.

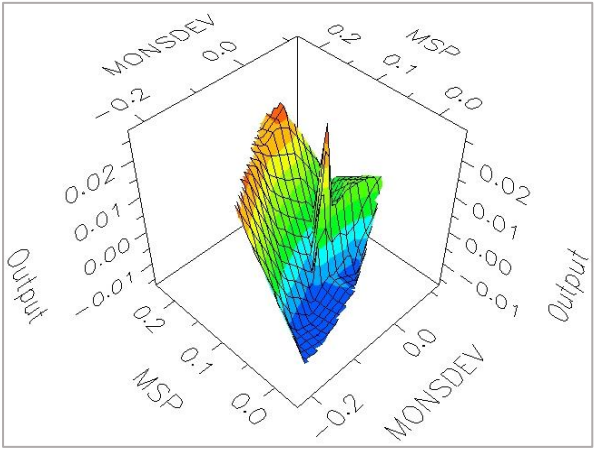

**(a)** Partial dependence plot for MSP and MonsDev

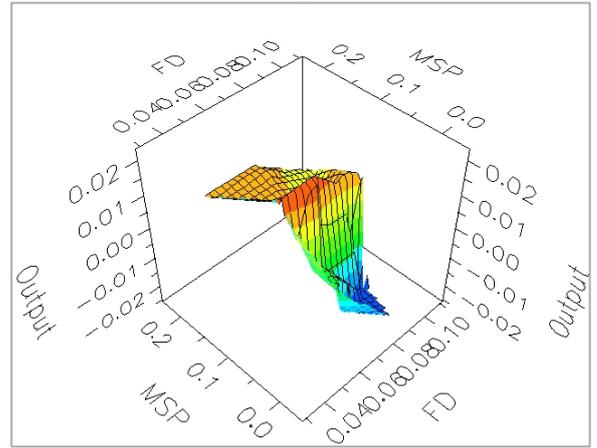

**(c)** Partial dependence plot for MSP and FD

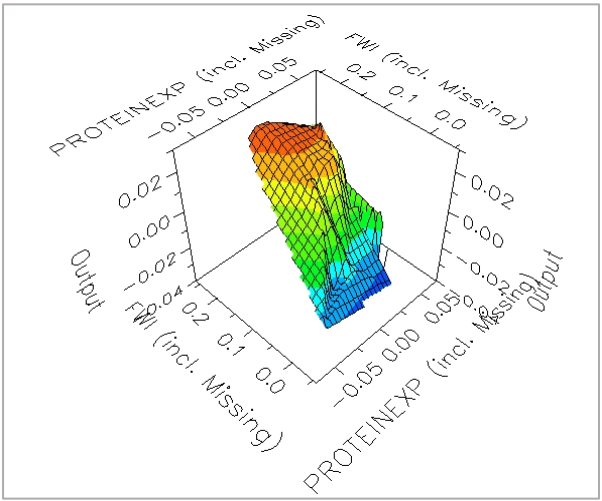

**(b)** Partial dependence plot for FWI and ProteinExp

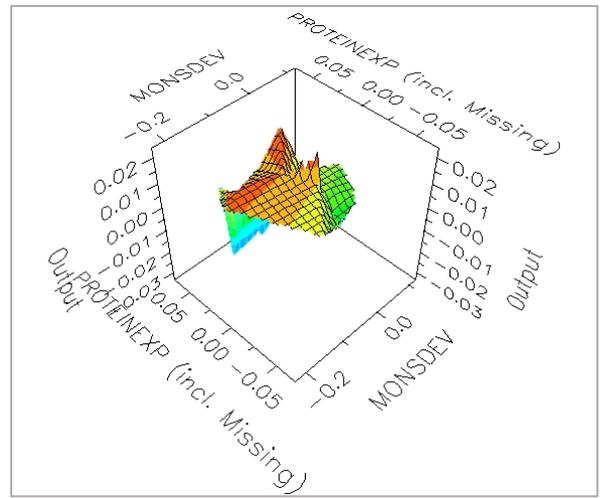

**(d)** Partial dependence plot for ProteinExp and MonsDev

**Figure 26.** Bivariate partial dependence plots for top interacting pairs

# 5. Policy Implications - The Way Forward

The challenge is to ensure food and nutritional security for growing Indian population with rising incomes as land and water resources continue to become more scarce. The demand-pull inflationary pressure on food prices is expected to increase as the economic growth picks up in the coming years. The obvious answer lies in raising the agricultural productivity and meet international standards through adoption of advanced agronomic technologies and investment in sustainable farming practices. However, the growing dependence on productivity for raising production to meet domestic demand might not help in easing inflationary pressures in the short to medium term and may cause prices to rise even more as introduction of new technologies increases the average cost of production during adoption years. This might lead to a situation where inflation in food prices would sustain in a period of rising agricultural output. That said, India certainly possess the ability to emerge as a multi-product agricultural powerhouse backed by its diverse topography, climate and soil, and to do so, India need to invest massively in building dense agri-supply lines, advanced agro-processing capabilities and organized retailing. A synergetic partnership needs to be developed between public and private players. Full relaxation in FDI limits in organized retailing and marketing of food products announced in the Union Budget FY16-17 is certainly a step in the right direction. This partnership could be further strengthened by franchising local shops to act as extended outlets of organized retailers. The risk of hoarding and black marketing by these new and already existing private players should be addressed with establishment of a national market regulator for food commodities. The primary function of any market is efficient price-discovery and agricultural markets in India are marred with frequent price manipulations, excessive middlemen commissions and poor competitiveness. The recent dilution of APMC act with launch of NAM (National Agricultural Market) - a pan-India electronic trading portal aggregating APMCs and other market yards across states is a welcome move towards creating a unified national market for agricultural commodities. Similarly, several archaic laws need to be revisited to increase competitiveness in domestic markets, such as *Essential Commodities Act, 1955* which discourages farmers and traders from investing in cold storages and warehousing facilities. Market-based reforms would not only reduce the spread between wholesale and retail food prices but would also improve the distressed socio-economic condition of farmers in India.

The rising farm wages is a positive sign and going forward, the focus should certainly not be on stalling this rise but rather on bringing down the pay-productivity gap. High agricultural labour productivity is essential for supplementing the higher demand arising out of increase in wages. The desired productivity move is achievable through investment in mechanization and extension of farms. Laws prohibiting farm land-lease markets needs to be reformed so as to create agricultural plots of economically viable size. Intensive capital requirements of mechanization could be addressed by enabling village *panchayats* in leasing farm machines. A symbiotic partnership could be struck with expansion of MGNREGS in agriculture, wherein a part of the wage on farm is paid by the government.

The Indian government needs to set up a strong institutional forecasting mechanism that could send demand and price signals to the farming community before sowing season, giving farmers the scope to scale up the production in accordance with expected demand. Additionally, this proposed mechanism could act as an early warning system for FCI in order to prepare for supply shortfall through international trade. The existing food management policy needs some big reforms starting with uncoupling of competing objectives served by procurement at MSPs and implementation of a *dynamic buffer* policy - timely release of food grains from the central pool into the open market in a year of subpar production and increasing domestic procurements, to replenish depleted buffer stocks, only during years with surplus production. The stocks in central pool should certainly not be allowed to overshoot the prescribed norms, unless there is a forecast of food crisis ahead. Adhering to buffer stock norms might not only ease food inflation[34] but also certainly reduce the burden on exchequer of procuring, storing and maintaining excess stocks, thereby helping the cause of fiscal consolidation. A renewed Indian government's commitment to fiscal consolidation, with a central fiscal deficit target of 3.5 percent of the total GDP for FY17, is commendable and would certainly support the disinflation process going forward. However, the current mechanism of administered price setting via MSPs and government interventions, through subsidies and differentiated tariffs, distort markets to an extent of suboptimal price discovery, thereby hindering transmission of implemented monetary and fiscal policies in the real economy, particularly in rural areas.

---

[34] Anand et al. 2016 estimates that a reduction in cumulative buffer stock intake of rice and wheat by 15 mn MT and 20 mn MT respectively during FY07-FY13 would have caused food inflation to decrease by about half percentage points per year.

The efficacy of MSP hikes to act as production incentive becomes questionable in itself as with inflationary effects of cost-indexed MSP on farm input costs and subsidies on the same farm inputs, the situation soon turns into *'a dog chasing its own tail'*. There is a need to evaluate whether an exclusive policy - providing price support for output or subsiding inputs - would be a sufficient stimulus for agricultural production. The principal role of MSPs should be the alignment of domestic prices along the long-term trend in international prices. Volatility and price spikes in global food prices is certainly not in the hands of any single nation, and a globally integrated Indian economy can make the most of it by developing a pro-active neutral trade policy (for both consumers and producers) along with a variable tariff structure, rather than outright bans on exports or imports.

The shift in dietary pattern toward pulses and other protein-rich items is certainly a welcoming sign, but to avoid a *'calorie catastrophe'*, India needs to develop and adopt sustainable farming techniques. There is a need to involve public agricultural research institutes and seed companies to develop short duration, high-yielding and pest-resistant varieties of pulses which are suitable for inter cropping and mixed cropping in arid conditions. Government needs to offer remunerative procurement prices for pulses, which would not only incentivise farmers with small un-irrigated plots but would also encourage cultivators with access to capital and irrigation to invest in pulses. The recently launched national mission (PMKSY) to expand cultivable area under irrigation and improve farm productivity is a welcome step forward in this regard. In addition to their nutritional advantage, pulses have low carbon and water footprints which make them essential to development of sustainable farming system. The increase in MSPs needs to complemented with procurement of pulses by FCI at the announced MSPs, similar to rice and wheat. The recent announcement by Indian government to create a buffer stock of 2 MT of pulses would certainly help farmers and act as a national protein security-net during the time of crisis. As a next step, the Indian government should explore inclusion of pulses in PDS for ensuring nutrition security to the poor.

The ability of Indian government to bring in major agricultural policy reforms and build synergetic investment partnerships with private players would not only determine the trajectory of food inflation in the coming years but also have a lasting effect on agricultural growth and ultimately, rural poverty rate.

## Supplementary Material

The dataset for replication of results presented in this study could be accessed at http://dx.doi.org/10.7910/DVN/JUNBRQ.